\documentclass[conference]{IEEEtran}
\IEEEoverridecommandlockouts
\usepackage{cite}
\usepackage{amsmath,amssymb,amsfonts}
\usepackage{algorithmic}
\usepackage{graphicx}
\usepackage{textcomp}
\usepackage{xcolor}

\usepackage[plain]{fancyref}
\usepackage{fancyhdr}

\def\BibTeX{{\rm B\kern-.05em{\sc i\kern-.025em b}\kern-.08em
    T\kern-.1667em\lower.7ex\hbox{E}\kern-.125emX}}
\begin{document}

\title{End-to-End Training of Neural Networks for Automotive Radar Interference Mitigation\\
\thanks{Research funded by Infineon Technologies AG and the Austrian Research Promotion Agency.}
}

\author{\IEEEauthorblockN{Christian Oswald}
\IEEEauthorblockA{\textit{Graz University of Technology}\\
Graz, Austria \\
christian.oswald@tugraz.at}
\and
\IEEEauthorblockN{Mate Toth}
\IEEEauthorblockA{\textit{Graz University of Technology}\\
Graz, Austria \\
mate.a.toth@tugraz.at}
\and
\IEEEauthorblockN{Paul Meissner}
\IEEEauthorblockA{\textit{Infineon Technologies AG}\\
Graz, Austria \\
paul.meissner@infineon.com}
\and
\IEEEauthorblockN{Franz Pernkopf}
\IEEEauthorblockA{\textit{Graz University of Technology}\\
Graz, Austria \\
pernkopf@tugraz.at}
}

\maketitle

\begin{abstract}
In this paper we propose a new method for training neural networks (NNs) for frequency modulated continuous wave (FMCW) radar mutual interference mitigation. Instead of training NNs to regress from interfered to clean radar signals as in previous work, we train NNs directly on object detection maps. We do so by performing a continuous relaxation of the cell-averaging constant false alarm rate (CA-CFAR) peak detector, which is a well-established algorithm for object detection using radar. With this new training objective we are able to increase object detection performance by a large margin. Furthermore, we introduce separable convolution kernels to strongly reduce the number of parameters and computational complexity of convolutional NN architectures for radar applications. We validate our contributions with experiments on real-world measurement data and compare them against signal processing interference mitigation methods.

\end{abstract}

\begin{IEEEkeywords}
Neural networks, machine learning, FMCW radar, radar interference mitigation, end-to-end training, convolutional neural networks, separable convolutions, object detection
\end{IEEEkeywords}

\begin{figure}[htp]
\includegraphics[width=\columnwidth]{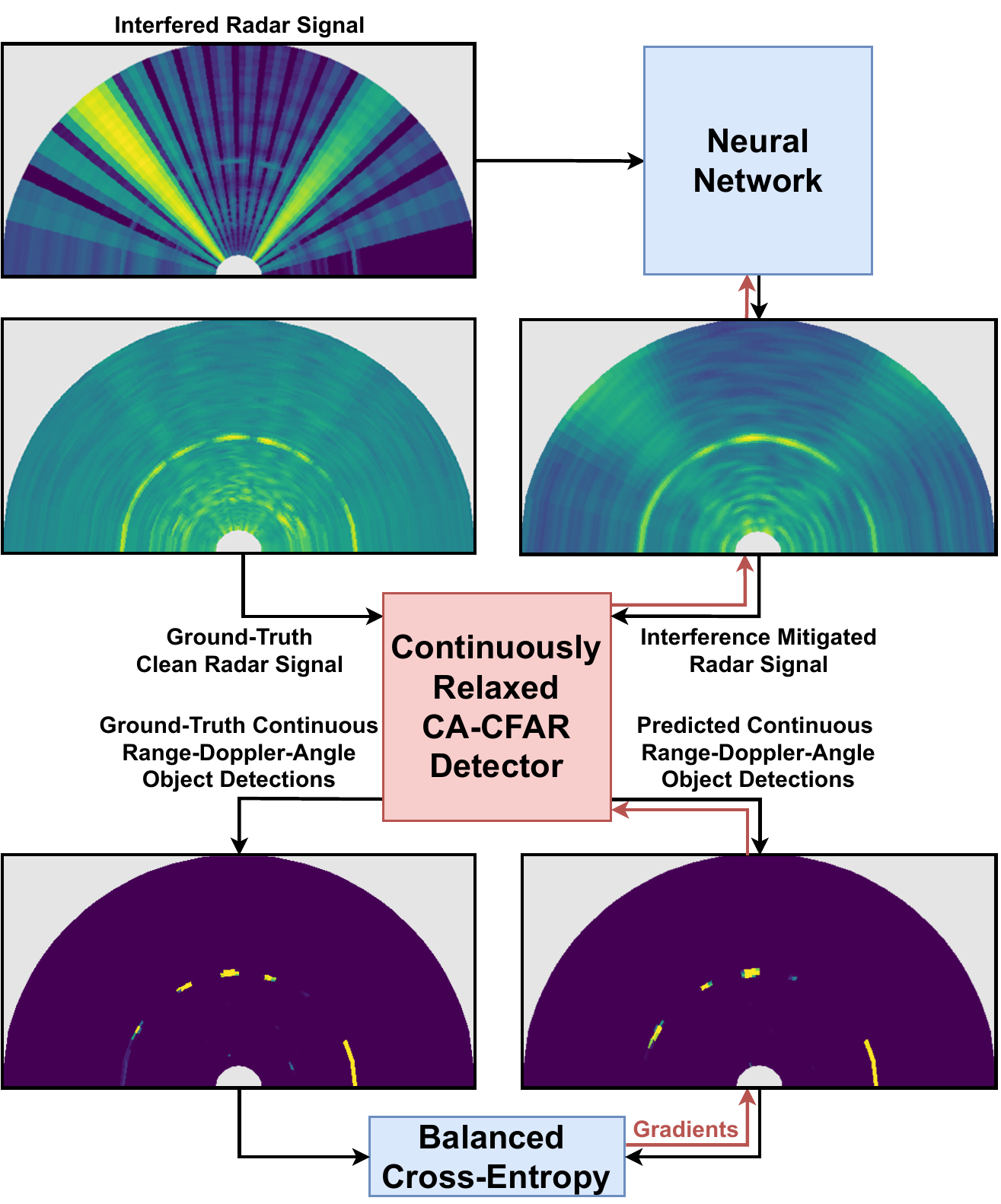}
\caption{Overview of the proposed method to train neural networks for radar interference mitigation. The network's output as well as the ground-truth clean radar signal are passed through a CA-CFAR detector, which has been designed prior to training. The CA-CFAR detector's threshold comparison is continuously relaxed during training, which yields continuous object detections and differentiability of the CA-CFAR detector. The predicted and ground-truth object detections are compared using the balanced cross-entropy loss function, and gradients are backpropagated through the CA-CFAR detector to the parameters of the neural network.} 
\label{fig:overview}
\end{figure}

\section{Introduction}
Frequency modulated continuous wave radar is a key technology in advanced driver assistance systems such as emergency break assist or adaptive cruise control. However, FMCW radar is prone to interference due to its emited wideband signal. More concretely, data delivered by an FMCW radar might be entirely corrupted if it is interfered by another FMCW radar (mounted on e.g., another car) operating in the same frequency range; This causes mutual interference as both radars are affected. A more detailed explanation of FMCW radar mutual interference can be found in \cite{brooker2007mutual, toth2018analytical}.

\subsection{Motivation and Contributions}
Many proposed methods for radar interference mitigation consist of NNs performing a regression from interfered to clean radar signals. However, as perfect interference mitigation is impossible in practice, residual errors remain in the neural networks' predictions. It is not clear how these residual errors influence object detection performance, which is the main application of automotive radar. In other words, there exists a misalignment between the neural networks' training objective and the actual desired objective, which prohibits usage of NNs in this safety critical application.

In this paper we therefore propose to train NNs directly on object detection maps. 
However, we do not integrate object detection into the neural networks' architecture, as this would increase their size while reducing their robustness and explainability. Instead, we continuously relax the well-established cell-averaging constant false alarm rate (CA-CFAR) detector \cite{scharf1991statistical} to enable end-to-end training of existing NN architectures. The NNs are therefore optimized for usage with a specific CA-CFAR detector. \Fref{fig:overview} depicts the proposed training setup including interference mitigation and object detection.

Another issue preventing NNs from deployment is their computational complexity. If NNs are to be implemented on hardware for real-time processing of radar data we must reduce their size as much as possible. In this paper we therefore present a simple yet effective modification for existing convolutional architectures, which reduces compuational complexity by orders of magnitude while not sacrifycing performance. In particular, we incorporate the independence of range, velocity and angle of objects into the NN architecture.  
\subsection{Related Work} \label{sec:related_work}
Mutual interference mitigation in FMCW radar is a well-studied problem. Some methods like frequency hopping \cite{bechter2016bats} change transmit parameters if interference is detected in the radar's output. Another approach, which also encompasses this paper, consists of removing interferences from an already corrupted signal. Zeroing \cite{fischer2016untersuchungen} detects interfered samples in the signal and sets them to zero. Ramp filtering \cite{wagner2018threshold} performs nonlinear filtering over multiple FM ramps to restore the magnitude of the object signal. Iterative method with adaptive thresholding (IMAT) \cite{marvasti2012sparse} reconstructs previously zeroed samples using the inverse FFT of the signal's sparse spectrum. 

There exist various machine learning based methods for mutual interference mitigation which operate on different signal representations:
\cite{mun2018deep}, \cite{mun2020automotive} process time-domain signals, while \cite{rock2021resource}, \cite{fuchs2021complex}, \cite{rock2020deep} perform interference mitigation on so-called range-Doppler (RD) maps, which are the time domain signal's two-dimensional Fourier transform. More concretely, \cite{rock2021resource} operates on one antenna at a time while \cite{fuchs2021complex} extends this to complex-valued array processing. Other architectures processing RD-maps are based on convolutional autoencoders \cite{fuchs2020automotive}, \cite{de2020deep}. In our previous work \cite{oswald2023angle}, we jointly process the entire receive array, but do so by using a fully convolutional architecture on range-Doppler-angle (RDA) maps, which is the three-dimensional Fourier transform of the receive array's output. The architecture proposed in \cite{ristea2020fully} transforms interfered signals given as spectrograms to interference mitigated range-profiles, which are the one-dimensional Fourier transform of the time domain signal. In \cite{wang2022prior} a fully convolutional architecture is applied to the interfered signal's short-time Fourier transform.

Deep learning has also been utilized to perform object detection in FMCW radar data. In \cite{major2019vehicle} an architecture using the single shot detector \cite{liu2016ssd} is trained; It operates on sequences of RDA-maps where labels have been annotated using a LiDAR sensor. The architecture in \cite{gao2020ramp} also trains on sequences of RDA-maps but is based on autoencoders. Reference \cite{brodeski2019deep} uses the radar calibration data to train a two-step object detector network in the style of faster-RCNN \cite{ren2015faster}. However, none of the aforementioned object detectors considers interference mitigation; Furthermore, they construct a fully trainable object detector while we do not replace the hand-designed CA-CFAR detector.
\section{End-to-End Training} \label{sec:ete}
 
\begin{figure}[tp]
\centerline{\includegraphics[width=5cm]{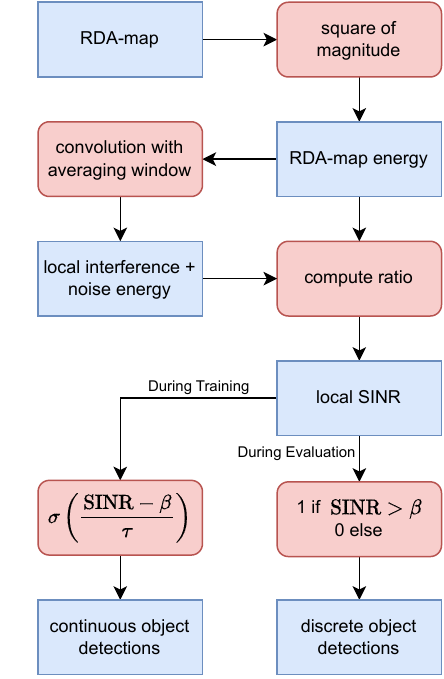}}
\caption{Computations performed inside a CA-CFAR detector. When training a NN, we replace the CA-CFAR detector's threshold comparison by the logistic sigmoid $\sigma$ to allow for backpropagation of gradients.
Operations are depicted in rounded and red, intermediate values in square and blue boxes.} \label{fig:cfar_structure}
\label{fig}
\end{figure}

\Fref{fig:cfar_structure} shows the structure of a CA-CFAR detector \cite{scharf1991statistical}. The key observation is that all operations inside a CA-CFAR detector except the threshold comparison are differentiable. We therefore can perform backpropagation through the CA-CFAR detector by replacing the threshold comparison with a differentiable surrogate function. The comparison operator can be continuously relaxed \cite{bengio2013estimating} using the logistic sigmoid 
\begin{equation}
    \sigma \! \left(\frac{\text{SINR}-\beta}{\tau}\right) = \frac{1}{1+e^{\frac{\beta-\text{SINR}}{\tau}}}, \label{eq:sigmoid}
\end{equation} 
where $\beta$ is the threshold value and $\tau$ the temperature hyperparameter which controls the smoothness of the continuous relaxation. If we let $\tau \rightarrow 0$, we recover the original discrete comparison operator, which is given as a step-function shifted by $\beta$. 

In this paper we assume the CA-CFAR detector's parameters, namely its window size, number of guard cells and threshold to be hand-designed.
Furthermore, we assume the CA-CFAR detector has a three-dimensional window for cell averaging. In other words, the CA-CFAR detector's takes RDA-maps as input and considers the vicinity of the cell-under-test w.r.t. range, velocity and angle to determine the local SINR. However, if we do not measure AoAs, we use a CA-CFAR detector operating on RD-maps with a two-dimensional window, and a four-dimensional CA-CFAR detector if we consider azimuth and elevation. Finally, we assume the CA-CFAR detector has a fixed, non-adaptive threshold. Depending on the NN architecture (cf. \Fref{sec:related_work}) we perform multiple Fourier transforms on the NN's output in order to feed range-Doppler-angle maps into the CA-CFAR detector, which are then included in the gradient backpropagation path.

The task at hand is a bin-wise binary classification between the two classes \textit{object} and \textit{no object}. However, these two classes are highly imbalanced as no object is present at the majority of bins in the RDA-map. Meanwhile, the commonly used binary cross-entropy (CE) loss function \cite{jadon2020survey} assumes the data set to be balanced, i.e., both classes appear with the same frequency in the data set. If we were to train a NN using CE, it would simply predict a low magnitude at all locations, reflecting the class frequencies in the data set. One common approach to address class imbalance consists of weighting the individual terms of the CE, resulting in the balanced cross entropy loss function (BCE) \cite{jadon2020survey},
\begin{equation}
    \text{BCE}(y, \hat{y}, \alpha) = -\alpha y \log (\hat{y}) - (1 - \alpha) (1-y) \log(1-\hat{y}), \label{eq:bCE}
\end{equation} 
where $y$ is the target, $\hat{y}$ the model's prediction, and $\alpha$ a weighting hyperparameter between $0$ and $1$. If $\alpha = 0.5$, both terms contributing to BCE are equally weighted, which is the desired weighting scheme for balanced datasets. If class $0$ appears more frequently in the data set than class $1$, one should set $\alpha > 0.5$: Misclassifying a sample of class $1$ then incurs a higher loss than misclassifying a sample of class $0$. 

A related technique to deal with non-differentiable operators in the computation graph is the so-called straight-through-estimator \cite{bengio2013estimating}, where operations prohibiting gradient backpropagation are replaced by a continuous relaxation in the backward path of the computation graph. However, in this work we also use the continuous relaxation \eqref{eq:sigmoid} 
in the forward path, i.e., we train on continuous targets $y \in [0, 1]$. Continuous targets ease optimization compared to discrete targets (which are used with the straight-through-estimator \cite{bengio2013estimating}) by providing more fine-grained feedback; Meanwhile, the discrete targets $y \in \{0, 1\}$ can be recovered by thresholding the continuous targets with $0.5$. Even though the BCE is most commonly used for discrete targets it extends to continuous targets as  
\begin{equation}
    \underset{\hat{y}}{\arg\min} (\text{BCE}(y, \hat{y}, \alpha)) = y, \; \forall y, \beta \in  [0, 1].
\end{equation}

\section{Separable Convolutions} \label{sec:separable}
Orthogonal to the training method described in \Fref{sec:ete}, we propose a method to reduce the overall complexity of convolutional NN architectures for radar interference mitigation by introducing separable convolution kernels.

A convolution kernel is called separable if it can be factored into multiple lower-dimensional kernels. For example, a two-dimensional kernel $W$ is separable if it can be written as $W = w_1 w_2$, where $W$ is a $K \times K$ matrix, while $w_1$ and $w_2$ have dimensions $K \times 1$ and $1 \times K$, respectively. When designing a convolutional NN one can therefore parameterize a convolution kernel by $w_1$ and $w_2$ instead of $W$. This reduces a kernel's number of parameters, as an $N$-dimensional kernel now only has $K \cdot N$ instead of $K^N$ parameters. 
On the other hand, the model is now restricted to learning separable filters (such as Gaussian or Sobel filters), which reduces performance in the general case. In radar interference mitigation however we can use separable convolutions to encode our domain knowledge about the independence of the range, velocity and angle of objects in the RDA-map. We can assume an object's range and angle to be independent if the distance between receivers is sufficiently small compared to the range of an object, i.e., the object is in the far-field and reflects plane-waves \cite{johnson1993array}. Furthermore, an object's measured velocity is independent from its range and angle if the object's relative movement doppler-shifting its reflection is negligible. From an image processing viewpoint, objects typically appear as axis-aligned star-shaped peaks in the RDA-map, where the length of the star's legs is given by the corresponding aperture of the radar sensor. These star shapes can also be described in a factorized form, such that convolutional NN architectures do not lose any of their expressiveness when scanning for this shape. 

Separable convolutions have lower computational complexity then generic convolutions with the same kernel size, as the input can be convolved with each of the separable kernel's components consecutively. The number of multiply-accumulate operations therefore reduces from $K^N$ \textit{multiplications} and $1$ \textit{accumulation} for generic to $K \cdot N$ \textit{multiplications} and $N$ \textit{accumulations} for separable kernels per convolution operation.

\section{Experimental Setup}
In \Fref{sec:experiments} we train our NN architecture introduced in \cite{oswald2023angle} on various training objectives and compare their performance with zeroing \cite{fischer2016untersuchungen}, ramp-filtering \cite{wagner2018threshold} and iterative adaptive thresholding (IMAT) \cite{marvasti2012sparse}. Moreover, we replace generic by separable convolutions and evaluate different kernel sizes. 

\begin{figure*}[htbp]
\centerline{\includegraphics[width=14cm]{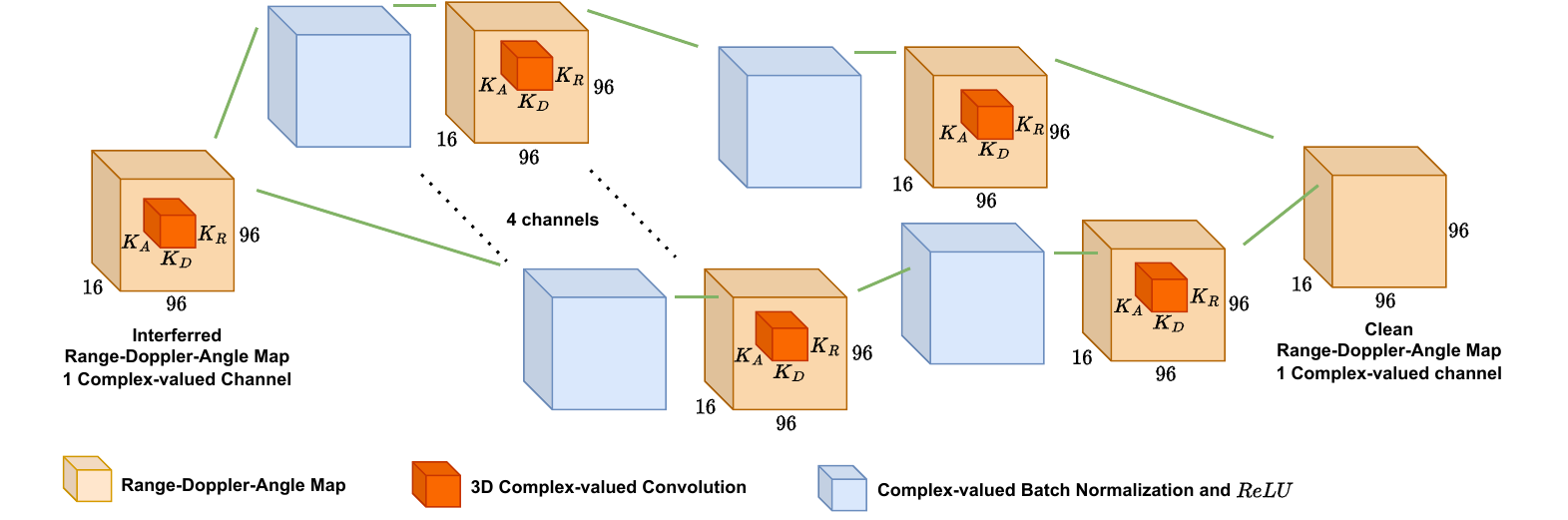}}
\caption{Structure of the Angle-Equivariant Neural Network (AENN) used for evaluation in \Fref{sec:experiments}. The network's input is a range-Doppler-angle map with dimensions $[96,96,16]$, which is processed with complex-valued convolution kernels of size $[K_A, K_D, K_R]$. The network has 2 hidden layers which are 4 and 2 channels wide. Note that AENN can be used with range-Doppler-angle maps of variable size due to its fully convolutional architecture. For a more detailed description, please refer to \cite{oswald2023angle}.} \label{fig:aenn_structure}
\label{fig}
\end{figure*}

\subsection{Angle-Equivariant Convolutional Neural Network (AENN)} \label{sec:AENN}
In this section we briefly review our architecture introduced in \cite{oswald2023angle} for FMCW radar mutual interference mitigation. We use AENN in subsequent experiments to validate our proposed end-to-end training setup as well as separable convolutions. A visualization of AENN can be found in \Fref{fig:aenn_structure}.

AENN consists of convolution, ReLU activation and batch normalization layers that operate on RDA-maps; All layers are complex-valued. The activations are zero-padded and the convolution stride is set to one such that AENN’s output (an interference mitigated RDA-map) has the same size as its input, which are interfered RDA-maps. The novelty of AENN is that it generalizes across different AoAs of interferences, i.e., only one AoA of interferences needs to be present in the training dataset; In other words, AENN is equivariant w.r.t. the AoA of interferences. If the ego radar measures azimuth and elevation, the three-dimensional layers can be extended by a fourth dimension to perform interference mitigation on range-Doppler-azimuth-elevation maps in an analogous manner.

The AENN we evaluate in subsequent experiments consists of three layers with 4, 2 and 1 complex-valued output channels, respectively. 
We use complex-valued $ReLU$ and batch-normalization with trainable scaling and bias \cite{fuchs2021complex} after the first and second layer. 

\subsection{Data Set}
Our data set consists of real-world inner-city measurements. 
We ran a CA-CFAR detector with a continuously relaxed threshold comparison on the measured RDA-maps to generate targets for AENN trained on BCE. For AENN trained on other objectives we directly use the measured RDA-maps as targets. We then synthesised and added artificial interferences to our measurements which are used as inputs for AENN. The interference signal model we use is the same as in \cite{brooker2007mutual, toth2018analytical, rock2021resource, wagner2018threshold}, amongst others. The data is finally divided into training, testing, and validation sets of sizes 2500, 250, and 250, respectively. A single sample comprises 96 range, 96 Doppler, and 16 angle bins, meaning one sample is a rank-three tensor with dimensions [96, 96, 16]. The ego radar's parameters as well as the parameter ranges for the artificial interferers can be found in the Appendix.

\subsection{Metrics and Loss Functions}

\subsubsection{Mean Squared Error (MSE)} 
To gauge the performance of AENN trained in the proposed end-to-end setup, we compare it against the same AENN trained to minimize the MSE between its prediction $\hat{S}$ and the clean RDA-map $S$, with
\begin{equation}
    \text{MSE}(S, \hat{S}) = \frac{1}{N} \underset{N}{\sum} (|S - \hat{S}|)^2, \label{eq:mse}
\end{equation} 
where $N$ is the total number of bins in a RDA-map.
\subsubsection{Mean Squared Error of Magnitudes (MAGMSE)}
As the CA-CFAR detector only operates on the magnitude of RDA-maps, one can also train NNs to minimize the MSE of RDA-map magnitudes. 
By contrast, when trained on \eqref{eq:mse}, NNs not only try to match the magnitude but also the phase of their prediction. We define MAGMSE as
\begin{equation}
    \text{MAGMSE}(S, \hat{S}) = \frac{1}{N} \underset{N}{\sum} (|S| - |\hat{S}|)^2.
\end{equation}
\subsubsection{F1-score} \label{sec:f1}
We use the F1-score to evaluate object detection performance. More concretely, we perform bin-wise comparison of ground-truth and predicted object detections, which are given as rank-three RDA binary tensors. The F1-score is defined as 
\begin{equation}
    F_1 = 2 \cdot \frac{N_{TP}}{N_{TP} + \frac{1}{2}(N_{FP} + N_{FN})},
\end{equation}
where $N_{TP}$ is the number of true positives, false positives $N_{FP}$ and false negatives $N_{FN}$. We evaluate the F1-score with a tolerance of 3 range bins, 3 Doppler bins and 1 angle bin; In other words, if a detected object has the aforementioned distance from a ground-truth object, we still count it as a true positive. 

\section{Experiments} \label{sec:experiments}

All AENNs have been optimized using ADAM and a batch-size of 8. We perform early-stopping w.r.t. the validation F1-score. We normalize the data set such that the covariance between the RDA-maps' real and imaginary part is the identity matrix. 
Through experimentation we found that we achieve best performance if we set the BCE's weighting parameter $\alpha = 0.75$ and the continuous relaxation's temperature $\tau = 10$. We hypothesize that higher values for $\tau$ lead to smoother gradients as predictions do not cap as quickly at $0$ and $1$. 
We compare various training objectives for AENN with and without separable convolution kernels. 
In \Fref{tab:aenn}, Generic convolutions are indicated by "g" and separable convolutions by "s". Furthermore, we evaluate multiple signal processing methods for interference mitigation. Note that each of the complex-valued convolution kernel's elements consists of two (real-valued) parameters. Furthermore, AENN performs complex-valued multiplications, which consist of four real-valued multiplications and two additions \cite{fuchs2021complex}.  

\begin{table}[htbp]
\caption{Performance of AENN.}
\begin{center}
\begin{tabular}{|c|c|c||c|c|c|c|}
\hline
\textbf{trained on} & \textbf{size} & \textbf{\#params} &\textbf{\textit{F1}} & \textbf{\textit{MAGMSE}} & \textbf{\textit{MSE}}\\
\hline \hline
BCE & [3,3,3] g & 800 & 0.844 & 129428 & - \\
\hline
BCE & [3,3,3] s & 296 & 0.867 & 154893 & - \\
\hline
BCE & [5,5,5] s & 464 & 0.903 & 152258 & - \\
\hline
BCE & [7,7,7] s & 632 & \textbf{0.921} & 148641 & - \\
\hline
BCE & [9,9,9] s & 800 & 0.914 & 151523 & - \\
\hline
\hline
MAGMSE & [3,3,3] g & 800 & 0.769 & 22192 & - \\
\hline
MAGMSE & [3,3,3] s & 296 & 0.748 & 23849 & - \\
\hline
MAGMSE & [5,5,5] s & 464 & \textbf{0.835} & 11600 & - \\
\hline
MAGMSE & [7,7,7] s & 632 & 0.827 & 11624 & - \\
\hline
MAGMSE & [9,9,9] s & 800 & 0.830 & 12066 & - \\
\hline
\hline
MSE & [3,3,3] g & 800 & \textbf{0.712} & 26438 & 34101 \\
\hline
MSE & [3,3,3] s & 296 & 0.621 & 53557 & 64813 \\
\hline
MSE & [5,5,5] s & 464 & 0.690 & 21311 & 30652  \\
\hline
MSE & [7,7,7] s & 632 & 0.628 & 39161 & 52641 \\
\hline
MSE & [9,9,9] s & 800 & 0.644 & 27368 & 38958 \\
\hline
\end{tabular}
\label{tab:aenn}
\end{center}

\vspace*{1em}
\caption{Performance of Reference Methods.}
\begin{center}
\begin{tabular}{|c||c|c|c|}
\hline
\textbf{Method} &\textbf{\textit{F1}}& \textbf{\textit{MAGMSE}}& \textbf{\textit{MSE}} \\
\hline
zeroing \cite{fischer2016untersuchungen} & 0.655 & 184397 & 207039 \\
\hline
ramp filtering \cite{wagner2018threshold} & 0.551 & 1275671 & 1327445 \\
\hline
IMAT \cite{marvasti2012sparse} & 0.513 & 297116 & 328553 \\
\hline
no mitigation & 0.306 & 12409976 & 12621186 \\
\hline
\end{tabular}
\label{tab:reference}
\end{center}
\vspace{-\baselineskip}
\vspace{-\baselineskip}

\end{table}

\begin{figure*}[htbp]
\centerline{\includegraphics[width=14cm]{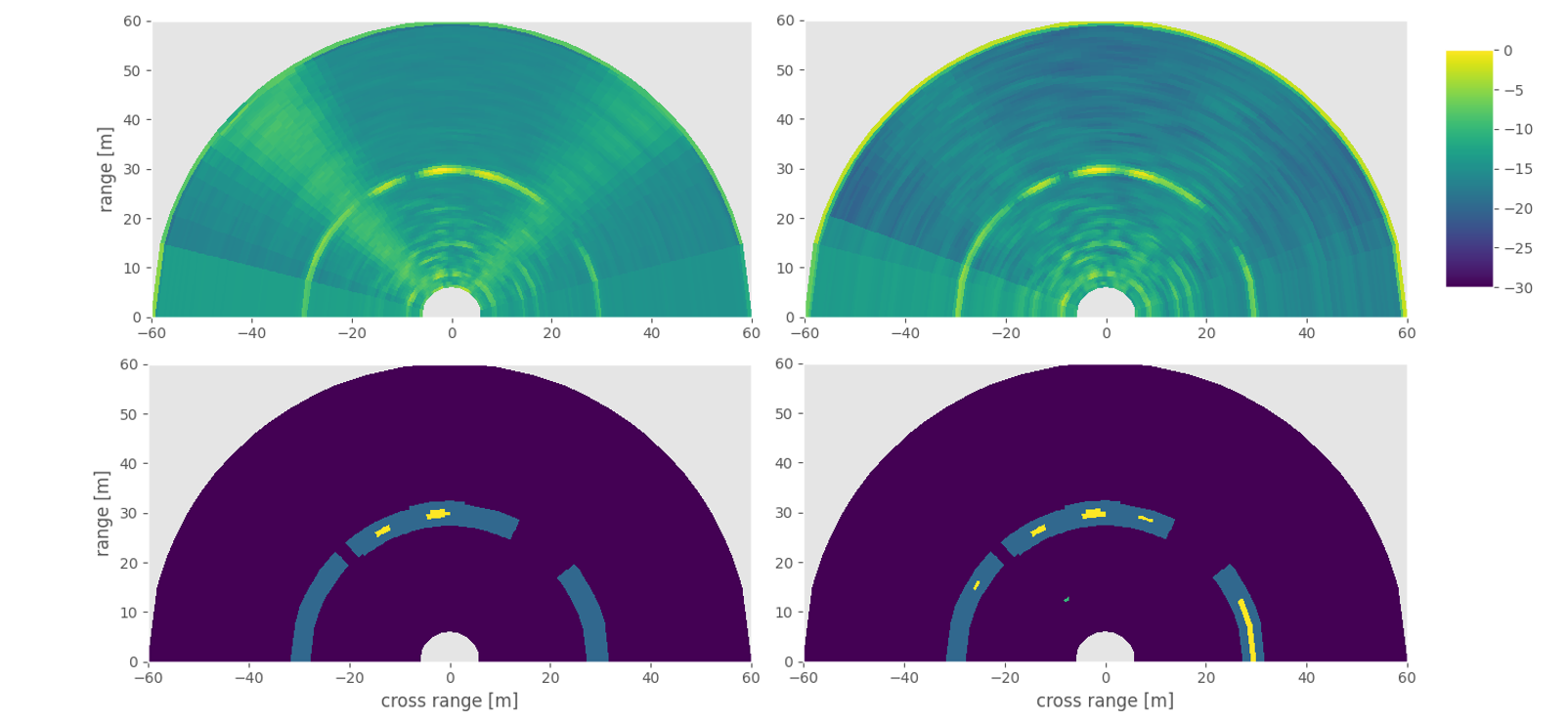}}
\caption{Comparison of an interference mitigated RDA-map (\textbf{top}) and its corresponding predicted objects (\textbf{bottom}) for the same AENN trained on the MSE of RDA-map magnitudes (\textbf{left}) and BCE of smoothened object predictions (\textbf{right}), visualized as range-angle maps. 
The objects are clearly visible in both range-angle maps; However, the AENN trained on MSE reduces its output's SINR such that the CA-CFAR misses many objects, while most objects are correctly detected when the AENN is trained on BCE. AENN pads its activations prior to convolution, which results in deviations of magnitudes at the maps' borders. The corresponding clean and interfered RDA-maps can be seen in \Fref{fig:overview}. The blue boxes in the detection maps indicate the tolerance for determining true positives as described in \Fref{sec:f1}. 
Yellow bins are therefore true positives, while the turquoise bin is a false positive. 
We performed a non-coherent sum over the Doppler-dimension of the RDA-maps and a sum of the object detections to arrive at these plots. The maps in the top row have been normalized such that their maximum values are zero dB. All plots have been up-sampled for better interpretability. Best viewed in color.}
\label{fig:example}
\end{figure*}

As can be seen in \Fref{tab:aenn}, training AENN on BCE outperforms other training objectives by a large margin in terms of the F1-score. Training AENN on MAGMSE instead of MSE already increases object detection performance. As expected, the model trained on MAGMSE achieves the lowest MAGMSE on the test set. An example of predicted RDA-maps and their associated object detections can be seen in \Fref{fig:example}. Even though the MAGMSE of AENN trained with BCE is multiple times higher than the MAGMSE of AENN trained with MAGMSE, the former's prediction actually looks more similar to ground truth, which can be found in \Fref{fig:overview}. This is due to the fact that AENN with MAGMSE smoothens the noise floor in addition to interference mitigation; This phenomenon has also been observed by \cite{rock2021resource}. 

The impact of separable convolutions on performance highly depends on the chosen training objective. While AENN with BCE greatly benefits from separable convolutions, performance of AENN with MSE actually deteriorates. This hints at the learned behaviour of AENN: When trained on BCE, AENN mainly performs template matching of the characteristic object peaks while suppressing everything else. As described in \Fref{sec:separable}, these object peaks can be represented in factorized form, which motivates the usage of separable convolutions. Meanwhile, restoring the clean complex-valued RDA-map is a more difficult task that seems to require generic convolutions. AENN trained on MAGMSE also benefits from separable convolutions, however, performance gains are not as pronounced as with BCE. Interestingly, performance of AENN trained with BCE increases when replacing generic with separable convolutions of same size even though AENN's expressivity is reduced; We hypothesize this improvement is a consequence of the reduced optimization space. 

AENN trained with BCE learns interference mitigation implicitly by uncovering masked objects. One therefore needs to ensure that a sufficient number of objects is masked by interference.
The MSE for AENNs trained with BCE and MAGMSE are not reported as it becomes arbitrarily high. Interestingly, zeroing performs best amongst classical intereference mitigation methods, as summarized in \Fref{tab:reference}. We suspect this is caused by the high level of interferences. For instance, ramp filtering only corrects the magnitude of an interfered signal, while corruptions proportional to the interference level remain in the signal's phase.

\section{Conclusion}
In this paper we introduced end-to-end training of neural networks for FMCW radar mutual interference mitigation. Furthermore, we applied separable convolutions to reduce the overall complexity of such NN architectures. These contributions address two main issues preventing NNs from being used for interference mitigation in practice, namely computation complexity and transparency of the learned behaviour. 
However, more work is necessary for NNs to be usable in practice; For instance, no guarantees can be provided that NNs behave as intended under all possible circumstances, e.g. drifting sensor characteristics or changing weather conditions. In other words, the robustness of such NNs needs to be increased. Furthermore, NNs for radar interference mitigation still lack explainability necessary for this safety critical application. Therefore, we devote future work to further improve the robustness and explainability of NNs for radar interference mitigation.

\bibliographystyle{IEEEtran}
\bibliography{IEEEabrv,conference_101719}

\section*{Appendix}
In \Fref{tab:ego} we list the ego radar's parameters we used in our measurement campaign. We generate artificial interferer signals by uniformly sampling from the values given in \Fref{tab:interferer}. Even though the Signal \& noise to interference ratio (SNIR) is positive in our data set, the \textit{local} SNIR might be negative, as e.g., visible in \Fref{fig:overview}.

\begin{table}[htbp]
\caption{Parameters of ego radar}
\begin{center}
\begin{tabular}{|c|c|}
\hline
\multicolumn{2}{|c|}{\textbf{Parameter}}  \\
\hline
Sweep starting frequency  & 79 \text{GHz} \\
\hline
Sweep bandwidth & 0.27 \text{GHz} \\
\hline
Sweep duration & 12,8 $\mu s$ \\
\hline
Number of sweeps & 128 \\
\hline
Number of receivers  & 16 \\
\hline
Window type (range \& Doppler) & Hann \\
\hline
Window type (angle) & Taylor \\
\hline
\end{tabular}
\label{tab:ego}
\end{center}

\vspace*{1em}
\caption{Parameter ranges for interference signals}
\begin{center}
\begin{tabular}{|c||c|c|}
\hline
\textbf{Parameter} &\textbf{\textit{minimum}}& \textbf{\textit{maximum}} \\
\hline
Number of interferers  & 1 & 3 \\
\hline
Sweep starting frequency  & 78.9 \text{GHz} & 79.1 \text{GHz} \\
\hline
Sweep bandwidth & 0.15 \text{GHz} & 0.25 \text{GHz} \\
\hline
Sweep duration & 12 $\mu s$ & 24 $\mu s$ \\
\hline
Number of sweeps & 100 & 156 \\
\hline
Angle of arrival  & $-90^{\circ}$ & $90^{\circ}$ \\
\hline
Signal \& noise to interference & 10 dB & 30 dB \\
\hline
\end{tabular}
\label{tab:interferer}
\end{center}
\end{table}
\vspace{-\baselineskip}

\end{document}